\documentclass[letterpaper, 10 pt, journal, twoside]{IEEEtran}

\IEEEoverridecommandlockouts                              

\usepackage{graphics} 
\usepackage{epsfig} 
\usepackage{amsmath} 
\usepackage{amssymb}  
\usepackage[linesnumbered,ruled,lined]{algorithm2e}
\usepackage{subcaption}
\usepackage[size=footnotesize]{caption}
\usepackage[hidelinks]{hyperref}
\usepackage{xcolor}
\usepackage{bm}
\usepackage{multirow}
\usepackage{makecell}

\newcommand{\etal}{\textit{et al. }}
\newcommand{\trsp}{{\scriptscriptstyle\top}}

\begin{document}
\title{Motion Mappings for Continuous\\Bilateral Teleoperation}%

\author{Xiao Gao$^{1,2}$, Jo\~ao Silv\'erio$^{2}$, Emmanuel Pignat$^{2}$, Sylvain Calinon$^{2}$, Miao Li$^{1}$ and Xiaohui Xiao$^{1,3}$
\thanks{Manuscript received: December, 11, 2020; Revised February, 25, 2021; Accepted March, 18, 2021; Date of
publication March 25, 2021.
This paper was recommended for publication by Editor-in-Chief A. Okamura and Editor J. Ryu upon evaluation of Reviewers' comments.
This work was supported by the China Scholarship Council (CSC, No. 201906270139), the LEARN-REAL project (CHIST-ERA, https://learn-real.eu/), the Swiss National Science Foundation, the Open Funding Project of National Key Laboratory of Human Factors Engineering (Grant NO. 6142222180311) and the Open Funding Project of National Key Laboratory of Science and Technology on Space Intelligent Control Laboratory (Grant NO. 6142208180301). \emph{(Corresponding author: Xiaohui Xiao.)}}
\thanks{$^{1}$Hubei Key Laboratory of Waterjet Theory and New Technology, Wuhan University, 430072 Wuhan, China (xiaogao@whu.edu.cn, miao.li@whu.edu.cn, xhxiao@whu.edu.cn).}%
\thanks{$^{2}$Idiap Research Institute, CH-1920 Martigny, Switzerland (e-mail: joao.silverio@idiap.ch, emmanuel.pignat@epfl.ch, sylvain.calinon@idiap.ch).}%
\thanks{$^{3}$National Key Laboratory of Human Factors Engineering, China Astronauts Research and Training Center, 100094 Beijing, China.}%
\thanks{Digital Object Identifier (DOI): 0.1109/LRA.2021.3068924}%
}%

\markboth{IEEE Robotics and Automation Letters. Preprint Version. Accepted March, 2021}
{Gao \MakeLowercase{\textit{et al.}}: Motion Mappings for Continuous Bilateral Teleoperation} 

\maketitle

\begin{abstract}

Mapping operator motions to a robot is a key problem in teleoperation. Due to differences between local and remote workspaces, such as object locations, it is particularly challenging to derive smooth motion mappings that fulfill different goals (e.g. picking objects with different poses on the two sides or passing through key points). Indeed, most state-of-the-art methods rely on mode switches, leading to a discontinuous, low-transparency experience. In this paper, we propose a unified formulation for position, orientation and velocity mappings based on the poses of objects of interest in the operator and robot workspaces. We apply it in the context of bilateral teleoperation. Two possible implementations to achieve the proposed mappings are studied: an iterative approach based on locally-weighted translations and rotations, and a neural network approach. Evaluations are conducted both in simulation and using two torque-controlled Franka Emika Panda robots. Our results show that, despite longer training time, the neural network approach provides faster mapping evaluations and lower interaction forces for the operator, which are crucial for continuous, real-time teleoperation.
\end{abstract}


\section{Introduction}
\label{sec::intro}

\begin{figure}[t]
  \centering
    \includegraphics[width=0.99\columnwidth]{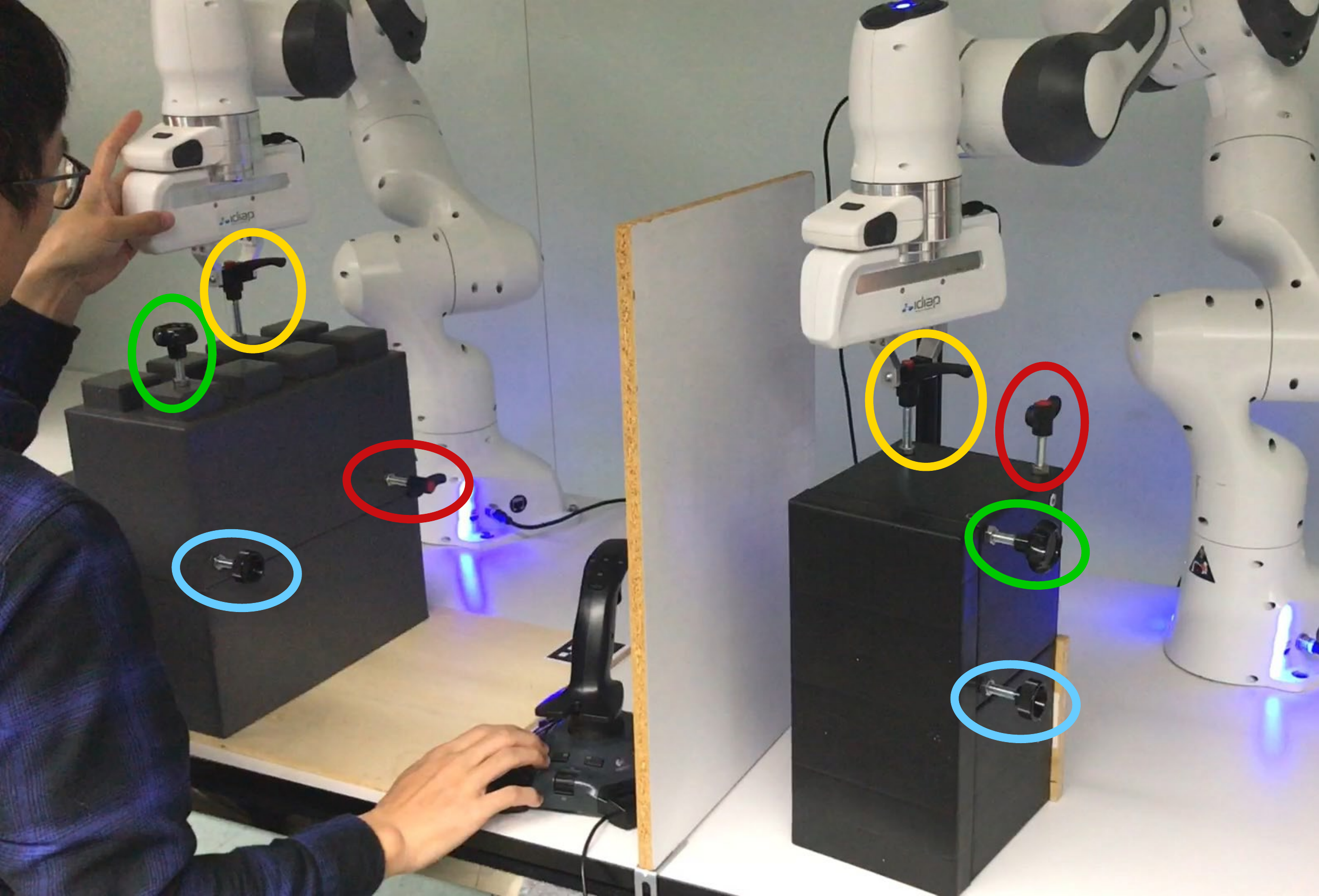}  
\caption{Valve turning experiment. The task consists of an operator kinesthetically guiding the local (left) robot to rotate four valves (in any preferred order) and the remote (right) robot executing the same behavior on its side, despite different valve poses. We consider four different valves on each workspace, where each pair has different poses with respect to the robot base (see colored circles). A vertical board prevents the user from relying on visual feedback to complete the task, emulating the realistic teleoperation of a remote robot.}
  \label{experiment_setup}
\end{figure}

\IEEEPARstart{D}{riven} by the need to perform tasks in remote environments, teleoperation has emerged as a central robotic paradigm with applications ranging from deep sea \cite{Murphy11} to outer space \cite{Lii10}. 
In this paper, we propose an object-centered motion mapping framework for continuous bilateral teleoperation. Our main objective is to address the problem of generating smooth, goal-directed robot trajectories when discrepancies exist between operator and remote robot workspaces. This problem, illustrated in Fig.~\ref{experiment_setup} with a valve turning example, remains largely unaddressed among the state-of-the-art teleoperation frameworks. 

\begin{figure*}
    \centering
    \includegraphics[width=0.99\textwidth]{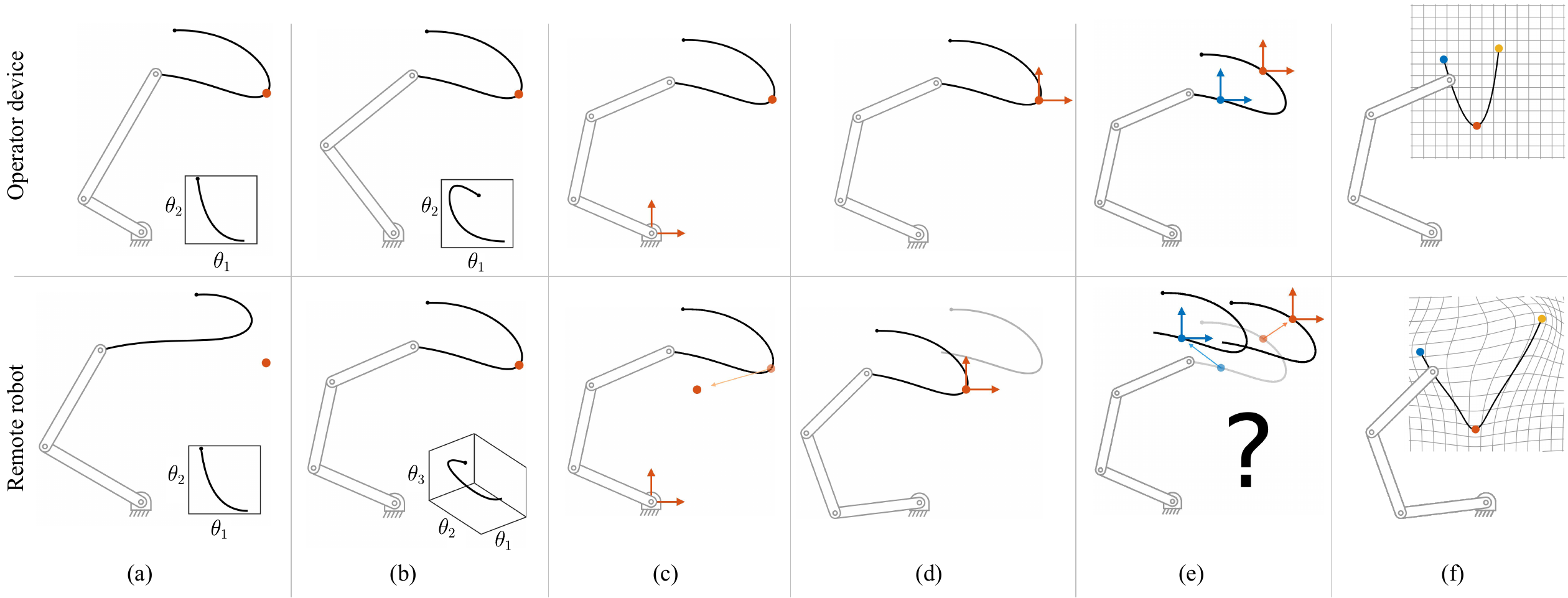}
    \caption{
      \textbf{Top row:} Operator side (here referred to as \emph{local environment}).
      \textbf{Bottom row:} Robot side (potentially a remote location).
      \textbf{(a)} Teleoperating a robot to pass through a point in space (orange) using a mapping in \textit{joint space} does not allow to handle kinematic differences (e.g., different link lengths or number of joints) between the teleoperating device and the remote robot.
      \textbf{(b)} A direct mapping in \textit{task space} can overcome kinematic differences.
      \textbf{(c)} However, it is not robust to differences in the environment (e.g. different object locations between workspaces).   
      \textbf{(d)} A possible solution is to formulate the tracking problem with a coordinate system attached to the object instead of the robot base. 
      \textbf{(e)} The question is: how to extend this strategy to multiple objects, while ensuring continuity, smoothness and precision?
      \textbf{(f)} Our proposed solution consists of finding an invertible and continuously differentiable mapping (also called a \textit{diffeomorphism}) that considers multiple objects (or other relevant landmarks in the workspace).
    }
    \label{fig:teleop}
  \end{figure*}
  
The logical steps between naive joint space mappings and object-centered representations can be seen in Fig.~\ref{fig:teleop}. 
Direct joint space mappings (Fig.~\ref{fig:teleop}(a)) impose strong embodiment constraints. Hence most teleoperation frameworks rely on task space representations as in Fig.~\ref{fig:teleop}(b). Despite that, even the most simple differences between local and remote environments can lead the remote robot to fail as in Fig.~\ref{fig:teleop}(c). This favours object-centered approaches (Fig.~\ref{fig:teleop}(d)) that are better but still tend to scale poorly with the number of objects (Fig.~\ref{fig:teleop}(e)). A common approach to this issue is to suspend the communication between the haptic device and the remote robot, relocate the haptic device and continue the task based on visual feedback \cite{conti2005spanning,scaled_teleoperation} (Section \ref{sec::related} will provide an overview). Nonetheless, the discontinuous manipulation leads to low efficiency and transparency, especially in tasks that involve multiple objects.
In this paper we propose an approach that reshapes the task space to ensure that the robot adapts to points of interest that can differ on the two sides (Fig.~\ref{fig:teleop}(f)).
Particularly, our main contribution is an object-centered formulation for task space motion mapping between the local and remote robot, with:
\begin{enumerate}
\item adaptability to different object poses between local and remote workspaces (either real or virtual);
\item invertible and continuously differentiable position and orientation mappings;
\item bilateral teleoperation capabilities, by relying on impedance control both for haptic guidance and compliant interaction of the robot with the environment.
\end{enumerate}

Our approach permits the operator to focus on its local setup (see Fig.~\ref{experiment_setup}), driving the input device to locally execute the task, with the remote robot performing it synchronously despite different object poses. We use a bilateral teleoperation setup with two Franka Emika Panda robots, where one acts as haptic device (on the operator side) and the other as a remote robot (operating on the targeted environment). We bring forward two possible implementations to achieve the proposed mappings: an iterative approach (Section \ref{sec::iteration}) and a neural network (NN) approach (Section \ref{sec::nn}). Experimental results (Section \ref{sec::experiment}) show that the latter is more advantageous due to faster mapping evaluations.


\section{Related Work}
\label{sec::related}

As the operator and remote robot workspaces are often different, a workspace motion mapping is required to compute the trajectory of the remote robot from the operator signals. The most common solution is position mapping \cite{conti2005spanning} by copying or scaling up/down the pose displacement \cite{scaled_teleoperation}. However, it requires the user to suspend the teleoperation and relocate the haptic device due to physical constraints, which results in a heavy workload and long time cost. To avoid this problem, rate control \cite{rate_control} is used by setting the velocity command of the remote robot proportional to the pose displacement of the local robot. But with this approach, it is hard to get accurate positioning. Workspace drift control \cite{conti2005spanning} can automatically center the physical workspace of the device towards the region of interest of the user. In \cite{liu2014modified,guanyang2019haptic,Mamdouh2012}, two methods, rate control for coarse motion and a variable scaled position mapping for accurate positioning, are combined to realize efficient and accurate teleoperation. However, for sequential teleoperation tasks, these methods still require a manual switching between the mapping modes. The teleoperation cannot be performed continuously. 
Moreover, they often heavily rely on visual feedback, which comes with additional limitations such as inferring 3D poses from 2D video feedback or the need to stream video, which can hinder applications where bandwidth is limited.

A relevant line of work relies on shared control approaches, which have been proposed, generally, as a means to reduce mental workload \cite{6212500} and improve performance \cite{haptic_virtual}. Trajectory distributions are used as virtual fixtures in \cite{Ewerton19}, which can constrain the robots to stay close to certain objects. Along these lines, Raiola \etal \cite{Raiola15} propose probabilistic virtual fixtures.
However, this approach requires demonstrations (as in \cite{Ewerton19}), and its generalization is not straightforward when the workspaces change. In the spirit of learning from demonstration, Zeestraten \etal \cite{Zeestraten18} propose an object-centered strategy for shared control, where the remote robot corrects the orientation of the end-effector to align with an object. Havoutis \etal \cite{havoutis2019learning} exploit a similar concept while considering multiple coordinate systems associated with different objects to resolve differences between local and remote environment configurations. With respect to this line of solutions, our approach does not rely on demonstrations and can ensure that moving towards an object on one side is replicated on the other.

Our solution takes inspiration from \cite{Perrin2016} to formulate the problem as that of computing a diffeomorphic mapping between workspaces. We begin by directly extending the iterative method proposed in \cite{Perrin2016} to fulfill our goals (Section \ref{sec::iteration}). But, as we show in Section \ref{sec::experiment}, despite fast training time, the time for updating the desired robot trajectory can be prohibitively high for real-time teleoperation. We build on this solution, and propose an approach based on a neural network that considerably alleviates the time constraints (Section \ref{sec::nn}). Additionally, we go beyond classical unilateral teleoperation and consider the more general case of bilateral teleoperation \cite{hokayem2006bilateral}. Indeed, we show that our approach is robust to haptic feedback originating from the remote side, which is leveraged to perceive how the teleoperated robot interacts with the environment. For the sake of stability, we rely on an impedance control implementation of bilateral teleoperation \cite{cho2005stable}, as opposed to using direct force feedback. 


\section{Problem Statement and Definitions}
\label{sec::problem}

In this section we propose a set of fundamental constraints that need to be fulfilled for robust bilateral teleoperation with different object poses. We then formalize them mathematically. Finally, orientation conversions and error definitions (which will be used in optimization) are introduced.

\subsection{Constraints for bilateral teleoperation with different object poses} 
\begin{figure}[t]
  \centering
    \includegraphics[width=0.45\textwidth]{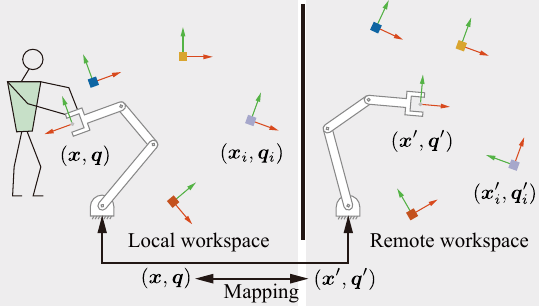} 
\caption{A schematic of bilateral teleoperation.
 \textbf{Left:} the local space built for teleoperation.
  \textbf{Right:} the remote workspace. Colored squares with arrows represent the poses of 4 objects to be manipulated. Our approach builds a mapping between the end-effector poses of local and remote robots.  
  }
  \label{tele_2D}
\end{figure}

We consider an arbitrary number of objects $N$ that are present in both workspaces following a one-to-one correspondence, as shown in Fig.~\ref{tele_2D}. The positions and orientations of the objects are represented as $\bm A = \{\bm x_{i}, \bm q_{i}\}_{i=1}^N$ and $\bm B = \{\bm x'_{i}, \bm q'_{i}\}_{i=1}^N $ respectively, where ${\bm x \in \mathbb{R}^D}$, ${\bm q \in \mathcal{S}^3}$. We consider cases where $N\ge 2$, $D = 2$ or $D = 3$. The current poses of the local and remote robot end-effectors are described as $(\bm x, \bm q) $ and $(\bm x', \bm q') $ respectively.
Our goal is to build a mapping $(\bm x, \bm q) \Leftrightarrow (\bm x', \bm q')$ between the two workspaces, where the following constraints apply:
\begin{enumerate}
  \item When the local robot approaches the $i$-th object, with $\bm x \to \bm  x_{i} $, the remote robot should also approach the corresponding object, with $\bm x' \to \bm  x'_{i} $.\label{constraint_pos}
   \item Relative orientation should be the same when manipulating objects, which means that if $\bm x = \bm x_i$,  $ \Delta \bm q_i = \bar{\bm q}_i * {\bm q} = \bar{\bm q}'_{i} * {\bm q}' $ where $\Delta \bm q_i$ is the orientation of the robot end-effector with respect to the $i$-th object\footnote{Here, $*$ denotes the product between two quaternions and $\bar{(\cdot)}$ denotes quaternion conjugation.}. \label{constraint_ori}
  \item Position and orientation mappings should be invertible and continuously differentiable.
  \item The gradient of the position mapping should be small enough so that the robot motion stays within safe velocity limits.
  \label{constraint_vel}
\end{enumerate} 

The first two constraints are aimed at ensuring object-directed motions on both sides with proper relative orientations. The goal of the third constraint is to guarantee smooth control commands. Firstly, the continuity of pose mappings ensures that a continuous trajectory of the local robot results in a continuous trajectory for the remote robot. Secondly, ensuring that pose mappings are continuously differentiable also guarantees that the resulting velocities are continuous. And thirdly, invertibility of the mappings is crucial in impedance-based bilateral teleoperation, 
since motions on the remote side should be uniquely mapped from and to the motion of the local side.
Finally, the last constraint sets a limit on the magnitude of the position mapping gradient to limit velocities and ensure safe operation.

\subsection{Mapping formalism}
Note that the constraints for position and orientation are formulated differently, even though they share the same goal of reaching a given object with the same relative pose on both sides. This is due to the fact that, when object orientations are known, applying an orientation offset close to the object is enough to guarantee correct relative orientations. Consequently, we can treat the position and orientation mapping problems separately by focusing on the position first. We formulate the position mapping problem as finding an invertible and continuously differentiable function, or mapping, between $\bm{x}$ and $\bm{x}'$. The forward and backward mappings $\bm\Phi_x$, $\bm\Phi^{-1}_x$, are defined as
\begin{equation}        
      {\bm x}' = \bm\Phi_x(\bm x) , \quad  {\bm x} = \bm\Phi_x^{-1}(\bm x').
      \label{pos_basic_mapping}      
\end{equation}

Since, by constraint \ref{constraint_ori}), the orientation offset depends on the current robot position, we propose to write the orientation mapping directly as a linear function, with
\begin{equation}
     {\bm q}' = \bm\Phi_q(\bm x,\bm q) = \bm g(\bm x) * \bm q, \quad  {\bm q} = \bar{\bm g}({\bm x})* \bm q'.
    \label{ori_basic_mapping}
\end{equation}

Here, $\bm\Phi_x$ and $\bm\Phi_q$ are invertible and continuously differentiable position and orientation mapping functions, and $\bm g(\bm x)\in\mathcal{S}^3$ is the difference between two end-effector orientations as a unit quaternion and as a function of position. 

From the proposed mappings \eqref{pos_basic_mapping}, \eqref{ori_basic_mapping}, velocity mappings are obtained trivially by computing partial derivatives:
\begin{subequations}
	\label{eq:velocities}
  \begin{align}        
     \dot{\bm x}' &=   \frac{\partial \bm\Phi_x(\bm x)}{\partial \bm x} \dot{\bm x}, \quad    \dot{\bm x} = \left( \frac{\partial \bm\Phi_x( \bm x)}{\partial {\bm x}}  \right)^{-1} \dot{\bm x}',     \\
     {\dot{\bm q}}' &=  \frac{\partial \bm g(\bm x)}{\partial \bm x} \dot{\bm x} * \bm q +\bm g(\bm x)* \dot{\bm q} ,\\
     {\dot{\bm q}} &=  \frac{\partial \bar{\bm g}({\bm x})}{\partial{ \bm x}} \dot{\bm x} * {\bm q}' + \bar{\bm g}(\bm x)*  \dot{\bm q}',
\end{align}
\end{subequations}
with resulting angular velocity computed as $\bm \omega = 2 \dot{\bm  q}* \bar{\bm q}$. Note that, as suggested by constraint \ref{constraint_vel}), the velocity mappings are scaled by the gradient of the position mapping.

Equations \eqref{pos_basic_mapping}--\eqref{eq:velocities} lay out the foundation of our approach. As we shall see in the following sections, continuity and continuous differentiability of the proposed mappings come naturally from the techniques we choose to represent $\bm\Phi_x$ and $\bm\Phi_q$. Note that, for the orientation mapping in \eqref{ori_basic_mapping}, there is a linear relationship between $\bm q'$ and $\bm q$. Thus invertibility is already fulfilled. We only need the function $\bm g(\bm x)$ to be continuously differentiable. Invertibility of the position mapping, however, requires an additional formal constraint to be fulfilled. According to the inverse function theorem, a continuously differentiable function $\bm\Phi_x(\bm x) $ is invertible if ${\mathrm{det}\left( \frac{\partial \bm\Phi_x(\bm x)}{\partial \bm x}\right)>0, \forall \bm x \in \mathbb{R}^D}$. With this in mind, we will impose constraints on the used representations such that invertibility is ensured.

\subsection{Orientation conversions}

In our approach we rely on unit quaternions due to their compactness and easy composition of orientations. Nonetheless, for some operations, we will need to represent orientations as a 3-dimensional Euclidean vector. For this,
we here recall the definition of the logarithmic function ${\log: \mathcal{S}^3 \mapsto \mathbb{R}^3}$  to convert unit quaternions accordingly \cite{Zeestraten2017}:
\begin{equation}
    \bm r = \text{log}(\bm q)=\left\{
\begin{aligned}
&\text{arccos}^{*}(q_v) \frac{\bm{q}_u}{\|\bm{q}_u\|},  & \mathrm{if} \>\> \|\bm{q}_u \|\ne 0\\
&[0, 0, 0]^{\trsp},  & \mathrm{otherwise} \\
\end{aligned}
\right. ,
\label{qua_log}
\end{equation}
where $\bm q = \left[q_v, \bm q^\trsp_u\right]^\trsp \in \mathcal{S}^3$, and $q_v\in\mathbb{R}$, $\bm{q}_u\in\mathbb{R}^3$ refer to the real and vector parts of the quaternion. arccos* is a modified version of the arc-cosine function, see \cite{Zeestraten2017} for details.
When required during training and reproduction, unit quaternions are retrieved using the exponential function:
\begin{equation}
    \exp(\bm r) = \left\{
\begin{aligned}
&\left[\cos(\|\bm r \|), \sin(\|\bm r\|) \frac {\bm r^\trsp}{\|\bm r\|}\right]^\trsp, & \!\!\!\mathrm{if} \>\> \|\bm r\| \ne 0\\
&[1,0,0,0]^\trsp, & \!\!\!\mathrm{otherwise.}
\end{aligned}
\right. 
\label{qua_exp}
\end{equation}

\subsection{Error definitions}

Based on constraint \ref{constraint_pos}) and \ref{constraint_ori}),
the mapping errors should be small enough to fulfill  task accuracy requirements. Position and orientation errors of the mappings are defined as
\begin{subequations}  
    \begin{align}
        F_p(\Phi_x(\bm A),\bm B) &= \frac{1}{N} \sum_{i=1}^{N}\| \bm\Phi_x(\bm x_i) - \bm x_i'  \|, \label{position_cost}\\ 
        F_q(\Phi_q(\bm A),\bm B) &= \frac{1}{N}  \sum_{i=1}^{N}  d(\bm\Phi_q(\bm q_{i}),  \bm q_{i}') ,
        \label{orientation_cost}
    \end{align}
      \label{error}
\end{subequations}
  \begin{equation}
     \text{with    } \quad d(\bm q_a, \bm q_b) = \| \log(\bm q_a * \bar{\bm q}_b  ) \|,
    \label{orientation_distance}
  \end{equation}
where the position distance function $F_p$ is the Euclidean distance and the orientation distance function $F_q$ is a quaternion metric.


\section{Mapping by Iterative Method}
\label{sec::iteration}
A promising way to build the mappings \eqref{pos_basic_mapping}--\eqref{ori_basic_mapping} is to use a smooth diffeomorphism. For this, we extend the \textit{fast diffeomorphic matching} method in \cite{Perrin2016} by including orientations. In this section, we compute $\bm \Phi_x(\bm{x})$, $\bm \Phi_q(\bm{x})$ as a composition of $K$ locally weighted translations and rotations $\bm\phi_1(\bm{x}),\ldots,\bm\phi_K(\bm{x})$ and $\bm\varphi_1(\bm{x}),\ldots,\bm\varphi_K(\bm{x})$, respectively. And we explain how to find their optimal parameters such that the properties defined in Section \ref{sec::problem} are respected. 

\subsection{Locally-weighted translations and rotations}
Following \cite{Perrin2016}, we consider \emph{locally weighted translations}
\begin{subequations}
  \label{position_mapping}
  \begin{equation}
    \bm\phi(\bm x) = \bm x +  k_{\phi}(\bm x) \bm v_1,
    \label{eq:loc_w_trans}
  \end{equation}
  \begin{equation}
    \text{with}\quad   k_{\phi}(\bm x)= e^{-\rho_1^2 \|\bm x - \bm c_1  \|^2},
  \end{equation}
  \end{subequations}
where $\bm v_1 \in \mathbb{R}^D$ is the direction of the translation  and $k_{\phi}(\cdot)$ is a radial basis function (RBF) kernel with center $\bm c_1$ and width $\rho_1$. If $\mathrm{det}\left(\frac{\partial \bm\phi(\bm x)}{\partial \bm x} \right)>0$, $\forall  (\bm x, \bm v_1)\in \mathbb{R}^D \times \mathbb{R}^D$, then $\bm\phi(\bm x)$ is a diffeomorphism.  A sufficient condition is $0<\rho_1 < \rho_\text{max}(\bm v_1)=\frac{1}{\sqrt{2}\|\bm v_1\|} e^{\frac{1}{2}} $ (see \cite{Perrin2016} for the proof).

For the orientation mapping, \emph{locally weighted rotations} are applied as
\begin{subequations}  
  \begin{equation}
      \bm\varphi(\bm q) = \bm v_2 ^{k_{\varphi}(\bm x)} * \bm q,
  \end{equation}
  \begin{equation}	
  \text{with}\quad  k_{\varphi}(\bm {x}) = e^{-\rho_2^2 \| \bm {x} - \bm c_2   \|^2}.
   		\label{orientation_RBF}
\end{equation}  
  \label{orientation_mapping}
\end{subequations}
Here, $\bm v_2 \in \mathcal{S}^3$ is a unit quaternion representing a rotation direction. $\bm v_2^{k_{\varphi}(\bm x)} $ refers to spherical linear interpolation, which we implement using the functions \eqref{qua_log}--\eqref{qua_exp}, between the identity quaternion and $\bm v_2$ with the weight $k_\varphi(\bm x) \in (0,1]$. $k_\varphi(\cdot)$ is a RBF kernel with center $\bm c_2$ and width $\rho_2$. $\bm v_2 ^{k_{\varphi}(\bm x)}$ is analogous to the term $k_\phi(\bm{x})\bm{v}_1$ in \eqref{eq:loc_w_trans}, but in orientation.

\subsection{Algorithm description}
\label{sec::iterative_description}

Following \cite{Perrin2016} we minimize the errors \eqref{position_cost}--\eqref{orientation_cost} using an iterative method, described in Algorithm \ref{algorithm1}, which includes the extension to orientation. 
It employs three hyperparameters $\{\mu, \beta_1,\beta_2\}$ that regulate the degree of distortion in the resulting map. A complete description is beyond the scope of this paper, but the reader can refer to \cite{Perrin2016} for details. Note here that $\beta_2$ has the same effect as $\beta_1$ but for orientation.
The output $\bm{\rho}, \bm{C}, \bm{V}$ are the  parameters of the locally weighted translations/rotations in the final mappings $\bm\Phi_x$ and $\bm \Phi_q$. To calculate these parameters, $\bm Z = \{\bm Z_x, \bm Z_q\} =\{\bm z_i, \hat{\bm q}_{i}\}_{i=1}^N$ is initialized by $\bm A$ and the following steps are run at each iteration $j$ until the errors in \eqref{error} are small enough:
\begin{enumerate}
  \item find the index $m  \in \{1,\ldots,N\}$ where $\bm z_{i}$ has the maximum position distance to $\bm x_i'$; 
  \item find the index $n$ where $\hat{\bm q}_{i}$ and $\bm q'_{i}$ has the maximum orientation distance as measured by \eqref{orientation_distance};
  \item store $\bm c_{1,j}$, $\bm c_{2,j}$, $\bm v_{1,j}$ and $ \bm v_{2,j}$ for \eqref{position_mapping}--\eqref{orientation_mapping}.
  \item solve a bound-constrained minimization problem of $F_p(\bm \phi(\bm Z_x), \bm B)$ under $\rho_1 \in (0,\mu\rho_{\text{max}}( \bm v_1 ))$;
  \item solve a bound-constrained minimization problem of $F_{q}(\bm \varphi(\bm Z), \bm B)$ under $\rho_2 >0$;
	\item update poses $\bm Z$ by \eqref{position_mapping}--\eqref{orientation_mapping} as
		 $\bm Z_x = \bm\phi_j (\bm Z_x)$ and ${\bm Z_{q} = \bm\varphi_j(\bm Z) = \bm v_{2,j}^{k_{\varphi,j}(\bm Z_x)} * \bm Z_q}$.
\end{enumerate} 

The final mappings are given by
\begin{subequations} 
\begin{align}
  \label{eq:final_maps_pos}
  \bm\Phi_x(\bm x)&= \bm\phi_K\circ \cdots\circ\bm\phi_2\circ\bm\phi_1 (\bm x)                , \\
  \bm\Phi_q(\bm x, \bm q)&= \bm\varphi_K \circ\cdots\circ\bm\varphi_2\circ\bm\varphi_1 (\bm x,\bm q)  = \bm g(\bm x)* \bm q,
\label{eq:final_maps_ori}
\end{align}
\label{eq:final_maps}
\end{subequations}
where $\circ$ refers to function composition, and the associative law of quaternions is adopted to simplify $\bm\Phi_q$. So the forward mapping has the same form as \eqref{pos_basic_mapping}--\eqref{ori_basic_mapping}. Backward mapping is computed with Newton's method. According to the continuity of composite functions and chain rule, the pose mappings \eqref{eq:final_maps} are continuously differentiable.

 \begin{algorithm}
    \DontPrintSemicolon
    \caption{\textit{Fast diffeomorphic matching} extended to orientation
    }
    \label{algorithm1}
    \LinesNumbered
    \KwIn{$\bm A = \{\bm x_{i}, \bm q_{i}\}_{i=1}^N$, 
        $\bm B = \{\bm x'_{i}, \bm q'_{i}\}_{i=1}^N$,}
    \SetKwInOut{Parameter}{Parameters}
    \Parameter{$K \in \mathbb{N}^{+}, 0<\mu<1, 0<\beta_1,\beta_2 \leq 1 $,  }
    \KwOut{ $\bm \rho = \{\rho_{1,j}, \rho_{2,j}\}_{j=1}^K$, 
      $\bm C = \{\bm c_{1,j}, \bm c_{2,j}\}_{j=1}^K$,  
      $\bm V = \{\bm v_{1,j}, \bm v_{2,j}\}_{j=1}^K$,}
    \textbf{Initialize} $\bm Z = \{\bm Z_x, \bm Z_q \}=  \{\bm z_i, \hat{\bm q}_{i}\}_{i=1}^N, \bm Z :=\bm A $,   \newline
    \While {mapping errors \eqref{error} not small enough}{
      $m := \underset{i \in [1,...,N]}{\arg\max}(\|\bm z_i - \bm x_i' \|) $ \;
      $n := \underset{i \in [1,...,N]}{\arg\max}(d(\hat{\bm q}_{i}, \bm q_{i}')) $ \;      
      $\{\bm c_{1,j},\bm c_{2,j}\} := \{\bm z_{m} , \bm z_{n}\}$\;
      $\{\bm v_{1,j}, \bm v_{2,j} \}:= \{\beta_1 (\bm x_{m}' - \bm c_{1,j}), (\hat{\bm q}_{n} * \bar{\bm q}_{n}')^{\beta_2}   \}  $\;
      $\rho_{1,j} := \hspace{-4mm} \underset{\rho_1 \in (0,\mu\rho_{\text{max}}( \bm v_{1,j} ))}{\arg\min}\hspace{-5mm}(F_p(\bm \phi(\bm Z_x), \bm B)) $ \;
      $\rho_{2,j} := \underset{\rho_2>0}{\arg\min}(F_{q}(\bm \varphi(\bm Z), \bm B)) $ \;
      $\{\bm Z_x, \bm Z_q\}:= \{\bm \phi_j(\bm Z_x),  \bm \varphi_j(\bm Z) \}$\;  }
    \end{algorithm}   

\section{Neural Network Mapping}
\label{sec::nn}
In this section, a NN-based approach for finding $\bm\Phi_x(\bm{x})$ and $\bm\Phi_q(\bm{x})$ is proposed. Particularly, we leverage the use of normalizing flows \cite{kobyzev2020normalizing}, another popular approach for building invertible mappings. We adopt two NNs for position and orientation mappings: a real-valued non-volume preserving (Real NVP) model \cite{DinhSB2016} and a fully connected NN, respectively. Then the two NNs are combined together for the motion mapping.

\subsection{Position mapping}
We adopt the idea of normalizing flows, which use a composition of bijective transformations to build a mapping between an \textit{observed space} and a \textit{latent space}. Here we will consider these as the local and remote workspaces in the teleoperation setup and apply the mapping with $K$ transformations as \eqref{eq:final_maps_pos}.

Following the structure of Real NVP \cite{DinhSB2016}, several coupling layers are built and each layer is a simple bijection as an affine transformation. Given a $D$-dimensional input $\bm x$, the output $\bm x'$ of an affine coupling layer follows 
\begin{equation}
    \left\{
\begin{aligned}
    \bm x'_{1:d} & = \bm x_{1:d}, \\
    \bm x'_{d+1:D} & = \bm x_{d+1:D}\odot e^ {\bm s(\bm x_{1:d})   } + \bm t(\bm x_{1:d}) ,
\end{aligned}
\right. 
\label{eq:forward_transf}
\end{equation}
where $\bm s$ and $\bm t$ are convolutional networks for scale and translation. $\odot$ refers to element-wise product. Backward propagation is given by
\begin{equation}
    \left\{
\begin{aligned}
    \bm x_{1:d} & =\bm x'_{1:d},  \\
    \bm x_{d+1:D} & = ( \bm x'_{d+1:D}- \bm t(\bm x_{1:d})   )\odot e^{ -\bm s(\bm x_{1:d})   } .
\end{aligned}
\right.
\label{eq:back_transf}
\end{equation}
The Jacobian of the forward transformation \eqref{eq:forward_transf} is given by
\begin{equation}
    \bm J(\bm x) = \frac{\partial \bm \Phi_x(\bm x)}{\partial \bm  x} = \begin{bmatrix} \bm{I}_d & \bm{0}\\
    \frac{\partial \bm x'_{d+1:D}}{\partial \bm x_{1:d}}& \text{diag}\left(e^{\bm s(\bm x_{1:d})} \right) \end{bmatrix},
\end{equation}
which is triangular and the diagonal elements are positive ($\bm I_d$ is an identity matrix). Thus ${\mathrm{det}\left( \frac{\partial \bm \Phi_x(\bm x)}{\partial \bm x}\right)>0, \forall \bm x \in \mathbb{R}^3}$ and the mapping is invertible, as discussed in Section \ref{sec::problem}. Similarly to the proof of the iterative method, we know the position mapping is continuously differentiable.

The cost function to train the network is designed as
\begin{equation}
    Q_1 = F_p(\bm\Phi_x(\bm A),\bm B) + \lambda_1 \frac{1}{N_1} \sum_{i=1}^{N_1}|\bm J(\bm x_i^s) - \alpha\bm{I}_D | ,
\end{equation}
where the second term is a penalty for keeping the scaling factor of the mapping to fulfill the velocity constraint \ref{constraint_vel}), by making the Jacobian matrix $\bm J(\bm x_i^s)$ approach the identity matrix $\bm I_D$ with a scaling variable $\alpha>0$. 
Additionally, $\bm x^s_i \sim \mathcal N(\bm \mu_a,3\bm \Sigma_{a})$ are $N_1$ end-effector position samples on the local side, where $\bm\mu_a$, $\bm{\Sigma}_a$ are computed from the distribution of positions in $\bm A$. The scalar $\lambda_1$ is the weight of the penalty term in the cost. 

\subsection{Orientation mapping}

A fully connected NN is designed to approximate $\bm g(\bm x)$ in \eqref{ori_basic_mapping}. It is trained in Euclidean space by the logarithmic function of quaternions as \eqref{qua_log}.
We use a four-layer NN with 24 units per layer and \emph{tanh} as the activation function to build the mapping $\bm x \mapsto \bm r$, where $\bm{r} =  \frac{2}{\pi} \log(\bm q' * \bar{\bm{q}})$ and $\bm{g}(\bm x) = \exp( \frac{\pi}{2} \bm r)$. So, $\bm{g}(\bm x)$ is a continuously differentiable function. 

By the constraint \ref{constraint_ori}), the observed data used for training are  $\{\bm x_i \mapsto \frac{2}{\pi} \log(\bm q_i'*\bar{\bm q_i})\}_{i=1}^N$.
The cost function of the orientation mapping is written as
\begin{equation}
        Q_2  = \frac{1}{N}  \sum_{i=1}^{N} d_i   + \lambda_2 F_q (\bm{g}(\bm x_j^s), \bm q_I ) 
        + \lambda_3 \sum_{n} \|\bm w_n\|^2,
\end{equation}
where $d_i = d\left(\exp(\frac{\pi}{2} \bm r_i) *{\bm q_i}, \bm q_i' \right)$ is the orientation distance and $\bm r_i$ is the output of the NN when $\bm x = \bm x_i$.
The second term is used to keep the orientation offset close to the identity quaternion. Finally, the third term is the L2 regularization term to reduce overfitting, and $\bm w_n$ are the weights of the NN. The scalars $\lambda_2$, $\lambda_3$ weigh the contribution of the last two terms. 

\subsection{Combined network for pose mapping}

The two proposed NNs are combined together and trained using the Adam optimizer. The combined cost function is written as $Q = Q_1 + Q_2$.
 The observed data, sample input-output pairs are set as $ \{\bm{x}_i \mapsto \bm{x}_i' , \frac{2}{\pi} \log(\bm{q}_{i}' * \bar{\bm{q}}_{i} )\}_{i=1}^N $  for training. $\bm \lambda = \{\lambda_i\}_{i=1}^3$ are the weight factors chosen empirically. 
 The parameter $\lambda_1$ trades off between smoothness of the mapping (required for constraint \ref{constraint_vel}) and accuracy. Hence, the higher its value the less likely the system is to exhibit prohibitively high velocities, but it might come at the cost of lower accuracy at the mapping points. Similarly, $\lambda_2$ and $\lambda_3$ decide the orientation smoothness.
 The velocity mappings are computed trivially based on the partial derivatives of the NN and \eqref{eq:velocities}.


\section{Toy Example in 2D}
\label{sec::simulation}

\begin{figure}[t]
  \begin{subfigure}[b]{0.5\textwidth}
  \parbox[t]{2.5mm}{\rotatebox[origin=c,x=0.2cm, y=1.0cm]{90}{\textbf{Original}}}
          \begin{subfigure}[b]{0.45\textwidth}
            \centering $N_s = 1$
              \includegraphics[scale=1]{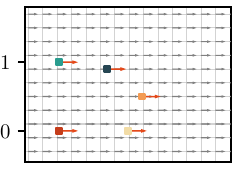}
          \end{subfigure}
          ~
          \begin{subfigure}[b]{0.45\textwidth}
       \centering $N_s = 100$
          \includegraphics[scale=1]{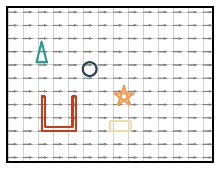}        
      \end{subfigure}
      \label{fig::2D_local}
  \end{subfigure}
  \parbox[t]{1.0mm}{\rotatebox[origin=c,x=0.2cm, y=1.0cm]{90}{\textbf{Iterative}}}
  \begin{subfigure}[b]{0.5\textwidth}
      \centering
      \begin{subfigure}[b]{0.45\textwidth}
      \centering
          \includegraphics[scale=1]{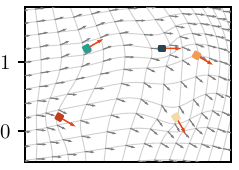}
      \end{subfigure} 
      ~
      \begin{subfigure}[b]{0.45\textwidth}
           \centering
          \includegraphics[scale=1]{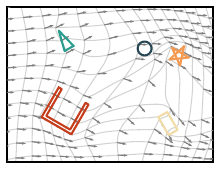}
      \end{subfigure}
      \label{fig::iterative_2D} 
  \end{subfigure}
  \parbox[t]{1.0mm}{\rotatebox[origin=c,x=0.2cm, y=1.4cm]{90}{\textbf{NN-based}}}
  \begin{subfigure}[b]{0.5\textwidth}
      \centering
      \begin{subfigure}[b]{0.45\textwidth}
           \centering    	
        \includegraphics[scale=1]{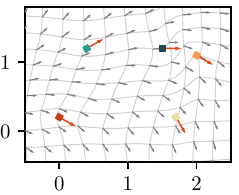}
      \end{subfigure} 
  ~
      \begin{subfigure}[b]{0.45\textwidth}
           \centering    
        \includegraphics[scale=1]{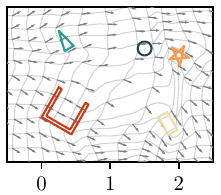}
      \end{subfigure}
      \label{fig::nn_2D} 
  \end{subfigure}
    \caption{Position and orientation mappings with the two methods, for $N_s = 1$ and $N_s = 100$.
  \textbf{1st row}: Local space (original).
  \textbf{2nd and 3rd rows}: Remote space distorted by the iterative method and NN-based method, respectively. 
    Five objects are located in both local and remote workspaces, and the gray/red arrows show the orientations of grid/key points. For better visualization, orientations are set as identity quaternions in the local spaces.
    } 
      \label{fig::2D_sim}
    \end{figure}

In this section, we compare the approaches from Sections~\ref{sec::iteration}--\ref{sec::nn} in a 2D scenario with 5 objects (Fig.~\ref{fig::2D_sim}). In order to test the robustness of our approach with respect to the number of points to be mapped, we consider three distinct cases, where the objects are represented using $N_s=1,10,100$ points.
For the iterative method, parameters are set as $\mu=0.6$, $\beta_1=0.95$, $\beta_2=0.9$. For the NN method, ${\bm{\lambda}=[0.002, 0.01, 0.3]}$, $N_1=200$ and $K = 4$. Both methods stop training when the errors in \eqref{error} are below 1 mm and 0.025 radians respectively. All tests were run in a system of a 2.8 Ghz Intel Core i7-7700HQ CPU (4 cores), 20 GB memory.
  
The results of estimating mappings on a 2D example are shown in Fig.~\ref{fig::2D_sim}. Grid points are plotted to visualize how the mappings distort the remote workspace. To get a clear visualization of orientation mapping, orientations of local workspace are simplified as identity quaternions (depicted as horizontal arrows). As only the orientation differences between the corresponding objects of two workspaces matter in \eqref{ori_basic_mapping}, this simplification is appropriate. Arrows are plotted to represent the x-axis direction.
We observe that the grid is reshaped smoothly by the proposed mappings. For the orientation mapping, the objects have almost the same directions as the nearby gray arrows, indicating that the computed mapping is correct and smooth around the points of interest. These observations apply to both methods.

Table \ref{tab_sim_compare} shows a comparison between the two approaches at the level of training and estimation time. The iterative method takes much shorter time for training than NN, but longer time for estimation. Because of using Newton's method to compute the backward mapping, it is more time-consuming than the forward mapping, which also happens for the velocity mapping. These preliminary results in 2D suggest that the time constraints can affect the performance of real-time robot control. On the other hand, the NN method shows a very efficient ability of estimation. Notice that the number of points used in the mapping increases the training time for both methods, and also the estimation time in the iterative method. This is because, as more points are added, more locally weighted translations and rotations are required to minimize the mapping errors. As a consequence, more functions need to be composed for evaluations. Notably, the NN-based approach is significantly more robust to the number of points in the mapping, taking the same evaluation time whether 5 or 500 points are used in total.

\begin{table}[t]
  \newcommand{\tabincell}[2]{\begin{tabular}{@  {}#1@{}}#2\end{tabular}}
  \renewcommand{\arraystretch}{1.3}
  \setlength{\abovecaptionskip}{-5pt}
  \caption{Comparison of simulation results.}
  \begin{center}   
  \begin{tabular}{cccc}
    \hline
    &$N_s$&\tabincell{c}{Iterative}&\tabincell{c} {NN-based} \\\hline
    \multirow{3}*{Training time (s)}   &1& \textbf{0.051} &${1.5 \times 10^{3}}$ \\
                                      ~&10& \textbf{0.61}   &${2.5 \times 10^{3}}$\\
                                      ~&100& \textbf{9.5}   &${4.6 \times 10^{3}}$\\\hline
    \multirow{3}*{Forward estimation (ms)}&1&  \textbf{0.44}   & 0.54 \\
                                        ~&10& 4.4       & \textbf{0.58}\\
                                        ~&100& 28      & \textbf{0.59}\\\hline
    \multirow{3}*{Backward estimation (ms)}& 1 & 1.2    & \textbf{0.55}    \\
                                          ~&10& 9.7     &\textbf{0.56}\\
                                          ~&100& 68   &\textbf{0.58}\\ \hline
    \multirow{3}*{Velocity estimation (ms)}&1  &  5.1    & \textbf{1.01}   \\ 
                                          ~&10&  45      & \textbf{1.09}\\
                                          ~&100& 320   & \textbf{1.04}\\\hline
  \end{tabular}
  \end{center}
  \label{tab_sim_compare}
  \end{table}


\section{Robot Experiments}
\label{sec::experiment}
 
\begin{figure*}[t]
  \vspace{0.3cm}
  \centering
  \begin{subfigure}[b]{0.23\textwidth}
    \includegraphics[width=\textwidth]{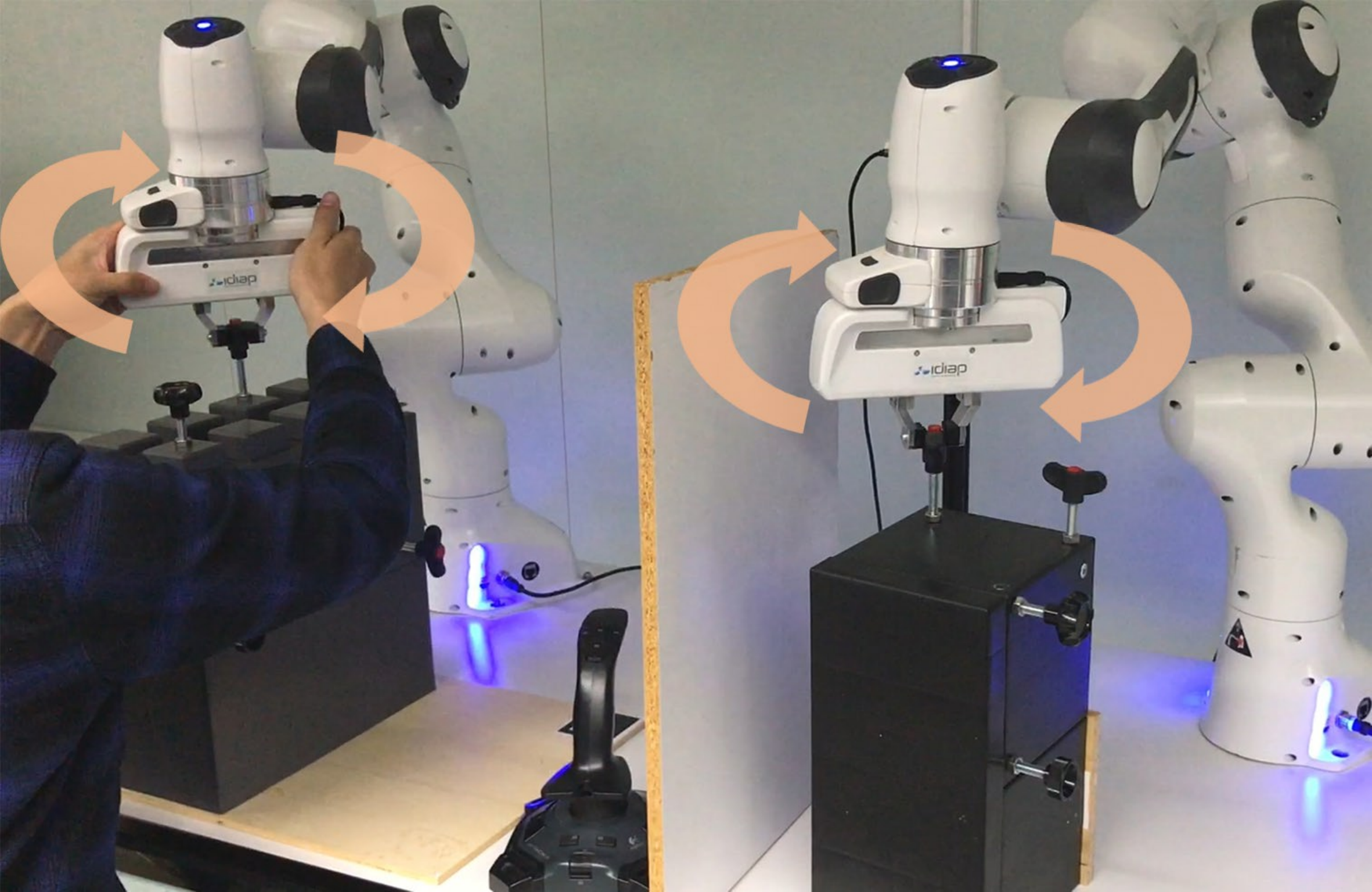}
    \caption{}
\end{subfigure}
~ 
  \begin{subfigure}[b]{0.23\textwidth}
    \includegraphics[width=\textwidth]{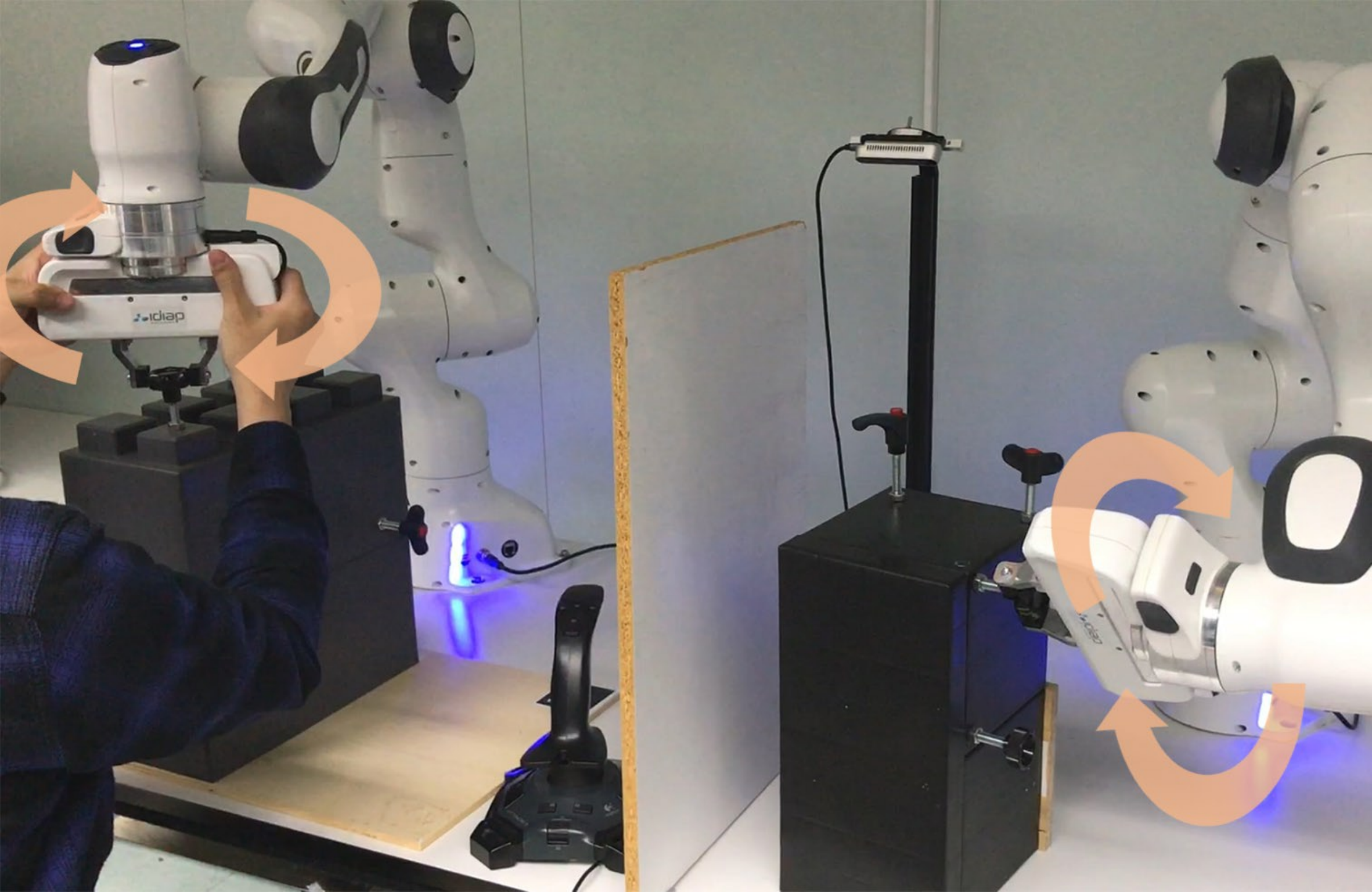}
    \caption{}
\end{subfigure} 
  ~ 
    \begin{subfigure}[b]{0.23\textwidth}
      \includegraphics[width=\textwidth]{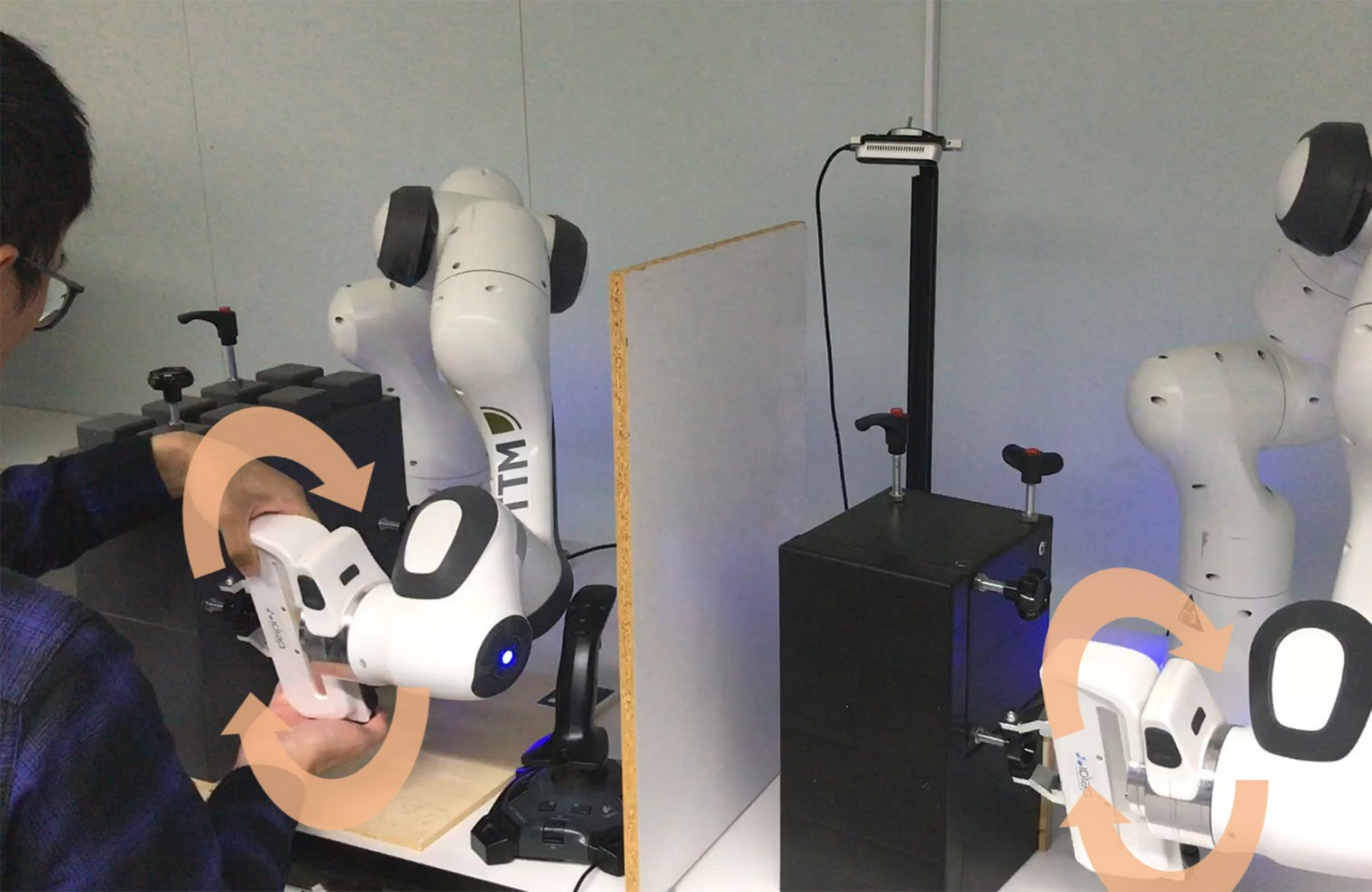}
      \caption{}
  \end{subfigure}
  ~ 
    \begin{subfigure}[b]{0.23\textwidth}
      \includegraphics[width=\textwidth]{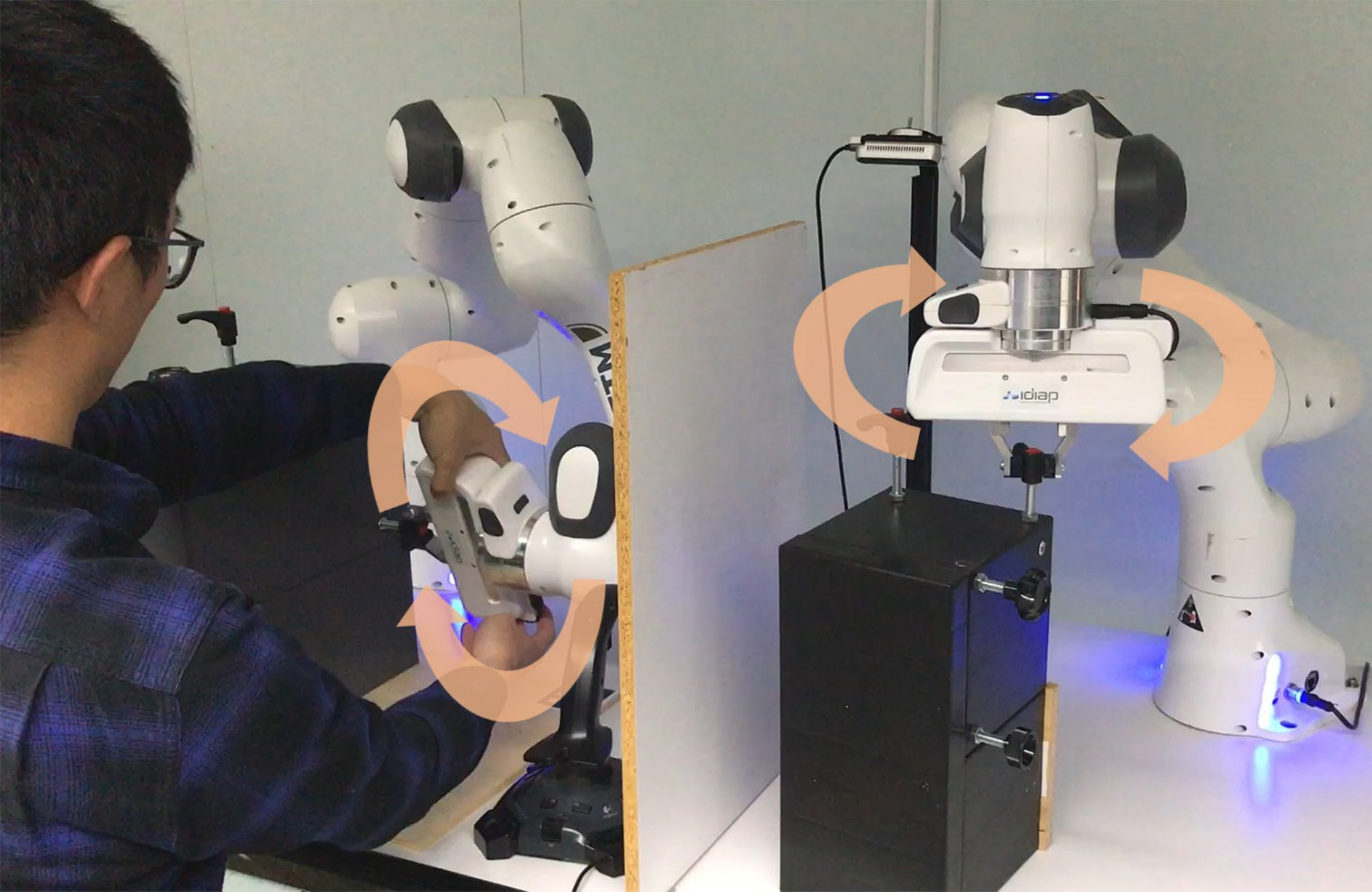}
      \caption{}
  \end{subfigure}
  \caption{Snapshots of the valve turning experiment. {(a)-(d)} show different valve poses on both workspaces and how the operator turns them without visual feedback.}
  \label{snapshots}
\end{figure*}

Valve-turning experiments were carried out to validate the two mapping methods in a realistic scenario, as shown in Fig.~\ref{experiment_setup}. Two torque-controlled Panda robots were used. The local robot is kinesthetically guided by the user, while the remote one executes the task in its surrounding environment by tracking reference poses generated by the proposed mappings. Four valves are attached to a block in both local and remote sides. The valve poses with respect to the robot bases differ between the two setups. A joystick is used to command the closing and opening of the grippers. An Intel Realsense D435 camera is mounted on the table to locate the blocks, and consequently the valves, by using two Aruco makers (the user does not use these video streams for teleoperation). The transformations between the valves and the markers, as well as between the camera and the robots, are calibrated in advance.

\subsection{Implementation}
Cartesian impedance control \cite{cho2005stable} is used on both robots, so that the user can guide the local robot and obtain haptic feedback from the interaction between the remote robot and the environment. The control law for both robots is  $\bm \tau = \bm M (\bm \theta)  {\ddot{\bm\theta}}  + \bm g(\bm \theta) + \bm c(\bm \theta,  {\dot{\bm\theta}}) +\bm J(\bm \theta)^\trsp \bm f$, with ${\bm f =\begin{bmatrix}\bm K_p (\bm x_d - \bm x) +\bm K_v(  \bm {\dot{x}}_d -  \bm {\dot{x}})  \\ 
	 \bm K_{pr} \log(\bm q_d* \bar{\bm q}) +\bm{K}_{vr}( \bm{\omega}_d -  \bm{\omega})  \end{bmatrix}},$
where $\bm \theta \in \mathbb{R}^7$ denote joint positions,  $\bm M(\bm \theta) \in  \mathbb{R}^{7 \times 7}$ is the mass matrix, $\bm c(\bm \theta,  {\dot{\bm \theta}}) \in \mathbb{R}^7$ is the Coriolis and centrifugal torque $\bm J(\bm \theta) \in  \mathbb{R}^{6 \times 7}$ is the Jacobian matrix, and $ \bm \tau$, $ \bm g(\bm \theta) \in \mathbb{R}^7$  are respectively control and gravitational joint torques. $\bm f \in \mathbb{R}^{6}  $ is the desired Cartesian force. $\bm x, \bm {\dot{x}} \in  \mathbb{R}^3 $  are the Cartesian position and velocity. $\bm q \in \mathcal{S}^3 $ is the unit quaternion and $\bm{\omega} \in \mathbb{R}^{3}$ is the angular velocity of the end-effector. $(\cdot)_d$ denotes the desired value, which is computed by the mappings in \eqref{pos_basic_mapping}--\eqref{eq:velocities}.  $\bm K_{(\cdot)}$ represents stiffness and damping gains (for position, orientation and velocity) of the Cartesian impedance controller.
The parameters of $\bm K$ can be different for the two robots. We introduce a scalar scaling factor to decrease the gains on the local robot such that interaction forces are lower and the robot is easier to guide by the operator. In this case, for the remote robot, $\bm K_p=600\bm{I}_3, \bm K_v=20\bm{I}_3, \bm K_{pr}=30\bm{I}_3, \bm K_{vr}=2\bm{I}_3 $. The scaling-down factor for $\bm K$ was set as $0.1$ for the local robot. The motor torque commands are sent at 1 kHz, and the reference trajectories are updated at 100 Hz. Parameters of the two methods are the same as Section \ref{sec::simulation}, and $N_s=1$.
\begin{figure}[t]
  \centering
  \begin{subfigure}[b]{0.5\textwidth}
    \centering
    \includegraphics[scale=0.48]{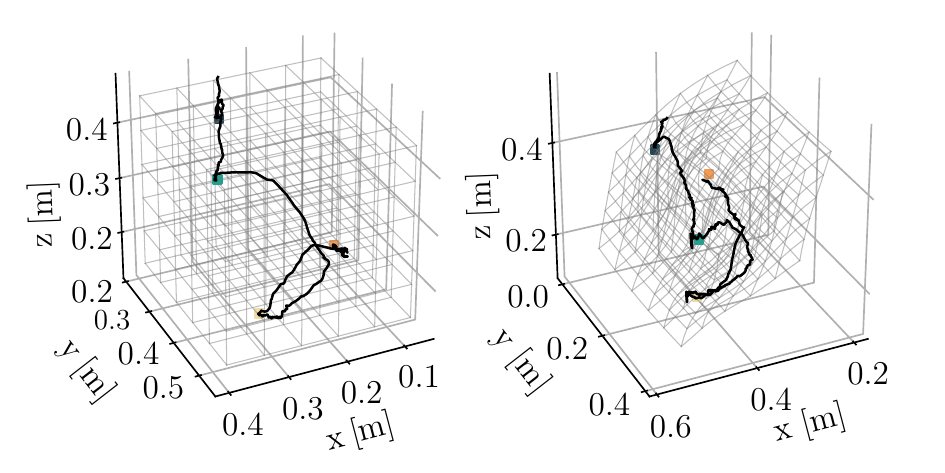}
    \caption{Mapping by iterative method}
    \label{iterative_3D}
\end{subfigure}
\begin{subfigure}[b]{0.5\textwidth}
  \centering
  \includegraphics[scale=0.48]{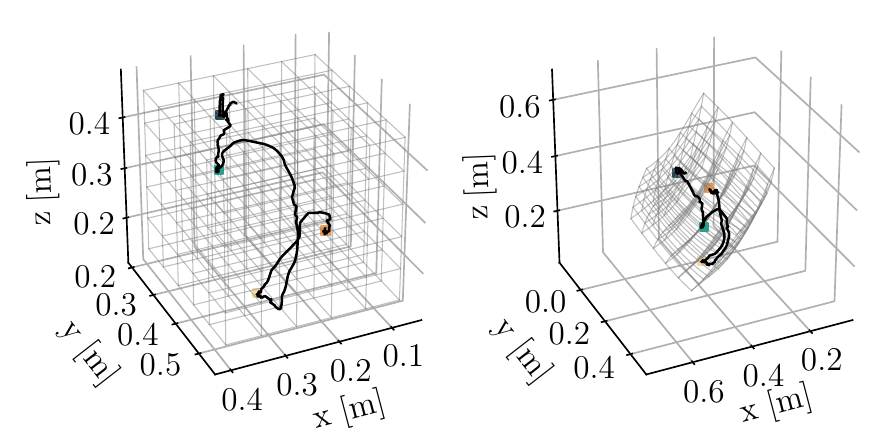}
  \vspace{-0.1cm}
  \caption{Mapping by NN method}
  \label{nn_3D}
\end{subfigure}
  \caption{3D workspace mapping with the two methods.
  \textbf{Left}: local workspace.
  \textbf{Right}: remote workspace. Robot trajectories are in black. Colored squares represent the position of the valves. Grid points are equally spaced on the left side. The distorted grid lines on the right side represent the corresponding mapped workspace. }
  \label{3D_figures}
\end{figure} 
\subsection{Experimental results}
The goal was to rotate four valves by $90^{\circ}$ in sequence. Figure~\ref{snapshots} shows snapshots of a reproduction. On the local side, valves were turned by human guidance one by one, and the remote robot performed the same task synchronously. The 3D trajectories of two robots in one trial are plotted in Fig.~\ref{3D_figures}. Similarly to the 2D case, cube grids and their respective distorted version are displayed to show the motion mappings. Although the valves on the two sides had different positions and orientations, both methods could deal with the differences, as the obtained mappings are smooth and generate trajectories that can adapt to the valve poses on both sides.

A traditional method of direct position mapping \cite{conti2005spanning} was used here as baseline to evaluate the two methods. The vertical board and valves of the left side were removed for the direct position mapping, so the user can directly watch the remote workspace, which emulates a typical approach where video streams are used.
Ten trials were held for each method. Table \ref{tab_real_compare} shows the results of the three methods. The iterative method takes longer than the NN method to turn all the valves and shows higher interaction forces applied by the operator. One plausible explanation is that the forward and backward estimation of the iterative method takes longer to compute than the NN method (see Table \ref{tab_sim_compare}), which brings a sensation of higher stiffness while guiding the robot (due to updating the poses at a lower rate). Both of them show better performance than the direct position method.
The NN method shows excellent ability for efficient forward and backward estimation, as shown by the lower estimation time in Table \ref{tab_sim_compare}, which results in a better transparency experienced by the operator (as demonstrated by the lower interaction forces).
One limitation is that the training time can be substantially higher than the iterative method for a totally new scenario. 
\begin{table}[t]
\newcommand{\tabincell}[2]{\begin{tabular}{@	{}#1@{}}#2\end{tabular}}
\renewcommand{\arraystretch}{1.3}
\caption{Comparison of experimental results.}
\begin{center}   
\begin{tabular}{cccc}
  \hline
    &\tabincell{c}{Iterative}&\tabincell{c} {NN-based} & \tabincell{c}{Direct} \\\hline
Task duration (s)      & $101 \pm 18$                 &         $\bm{96 \pm 16}$              &      $195 \pm  26.5 $ \\    
\tabincell{c}{Interaction force\\[-1mm] of the left robot (N) }  &  $1.12 \pm 0.21  $       &  $\bm{0.81 \pm 0.14 }$  & $1.05 \pm 0.15 $ \\
\hline
\end{tabular}
\end{center}
\label{tab_real_compare}
\end{table}


\section{Discussion}
\label{sec::discussion}

In this section, we discuss the limitations and additional considerations for the proposed approaches.
In the valve turning experiment, only the grasping poses are given as inputs to the mapping methods but one can potentially add further constraints. For example, viapoints can be added between two objects to bring additional geometric features or virtual guide priors, such as following a straight line between objects. Besides, if an object cannot be simplified as a single point for manipulation, more feature points can be added to describe objects. As shown in section \ref{sec::simulation}, the real-time performance of the neural network approach is not affected by the number of points in the mapping, making it an excellent candidate for scenarios that require a high number of points.

While the approach targets teleoperation in highly unstructured environments, it should be noted that, when feasible, it is good practice to design the local and remote workspaces to be as similar as possible. 
In extreme cases where the local and remote workspaces have significantly different dimensions, it is convenient to smoothen the distortions in the mappings by adjusting $\rho$ and $\beta$, in the case of the iterative mapping, and $\bm \lambda$, in the case of the NN approach. It should be kept in mind, however, that the precision of the mapping at the desired points might be affected.

Two impedance controllers were adopted here for bilateral teleoperation, providing indirect force feedback. Alternatively, other control methods can be used to implement bilateral teleoperation, such as direct force feedback. This should be done by taking into account that for some controllers, maintaining the stability of the closed-loop system in contact tasks might be difficult. Finally, in the event of considerable communication delays between the two robots, the mapping approach would still be applicable provided that the magnitude of the haptic feedback from the remote side would be decreased (possibly removed). This would correspond, in practice, to a more unilateral control approach.

Finally, due to the higher training time, the neural network approach might be prohibitively slow for cases in which the objects of interest move, which would require the network to be re-trained on-the-fly. We plan to further investigate this research challenge in future work. We also plan to extend the approach to the context of virtual fixtures.

The codes and the video of experiments are provided at \url{http://sites.google.com/view/bilateralmaps}.


\section{Conclusion}
\label{sec::conclusion}
We proposed a motion mapping framework for continuous bilateral teleoperation. The novel idea of using diffeomorphic mappings for this purpose is introduced, with two possible implementations being compared, one based on an extension of the iterative approach from \cite{Perrin2016} and an original one based on neural networks. We show that both can realize continuous bilateral teleoperation and generate an object-centered mapping, which is used to handle the differences between local and remote environment configurations. A 2D simulation has first been presented to visualize both position and orientation mappings. The results show that the NN-based solution provides faster estimation (about 2 ms) and is not affected by the number of points in the mapping, but needs a long time for training, while the iterative method can be trained very fast. We then validated both solutions in a valve-turning experiment using two torque-controlled Franka Emika Panda robots, that we contrasted to a direct position mapping as baseline \cite{conti2005spanning}. Our methods show better performance than the direct position mapping. The NN-based implementation shows lower interaction forces on the local side and lower task duration.



\bibliographystyle{IEEEtran}	
\bibliography{root.bbl}

\end{document}